\documentclass{article}

\usepackage{microtype}
\usepackage{graphicx}
\usepackage{subfigure}
\usepackage{booktabs} 
\usepackage{soul}
\usepackage{enumitem}
\usepackage{hyperref}

\usepackage[accepted]{mlsys2020}
\usepackage{multirow}
\usepackage[table,xcdraw]{xcolor}
\usepackage{diagbox}
\usepackage{comment}
\usepackage{tikz}
\newcommand*\circled[1]{\tikz[baseline=(char.base)]{
            \node[shape=circle,draw,inner sep=0pt] (char) {#1};}}

\newcommand{\ttb}{\linespread{0.8}\bf\footnotesize\ttfamily}
\newcommand{\ttm}{\linespread{0.8}\footnotesize\ttfamily}

\usepackage{color}
\definecolor{ballblue}{rgb}{0.13, 0.67, 0.8}
\definecolor{bleudefrance}{rgb}{0.19, 0.55, 0.91}
\definecolor{altred}{rgb}{0.98,0.14,0.56}
\definecolor{deepred}{rgb}{0.7,0,0}
\definecolor{deepgreen}{rgb}{0,0.5,0}
\definecolor{lightgreen}{rgb}{0,0.6,0}

\usepackage{listings}

\let\origthelstnumber\thelstnumber
\makeatletter

\newcommand*\Suppressnumber{%
  \lst@AddToHook{OnNewLine}{%
    \let\thelstnumber\relax%
     \advance\c@lstnumber-\@ne\relax%
    }%
}

\newcommand*\Reactivatenumber[1]{%
  \setcounter{lstnumber}{\numexpr#1-1\relax}
  \lst@AddToHook{OnNewLine}{%
   \let\thelstnumber\origthelstnumber%
   \refstepcounter{lstnumber}
  }%
}

\makeatletter
\let\orig@lstnumber=\thelstnumber

\newcommand\lstsetnumber[1]{\gdef\thelstnumber{#1}}
\newcommand\lstresetnumber{\global\let\thelstnumber=\orig@lstnumber}
\makeatother

\ifdefined\submit
\newcommand{\david}[1]{{}}
\newcommand{\kun}[1]{}
\newcommand{\sh}[1]{{}}
\newcommand{\mert}[1]{{}}
\newcommand{\jx}[1]{{}}
\newcommand{\dc}[1]{{}}
\newcommand{\wmh}[1]{}
\newcommand{\eiman}[1]{{}}

\else
\newcommand{\david}[1]{[{\color{orange}David: #1}]}
\newcommand{\kun}[1]{[{\color{blue}Kun: #1}]}

\newcommand{\sh}[1]{[{\color{purple}SH: #1}]}
\newcommand{\mert}[1]{[{\color{red}Mert: #1}]}
\newcommand{\jx}[1]{[{\color{red}JX: #1}]}
\newcommand{\dc}[1]{[{\color{red}DC: #1}]}
\newcommand{\wmh}[1]{[{\color{red}WMH: #1}]}
\newcommand{\eiman}[1]{[{\color{red}Eiman: #1}]}
\fi

\usepackage{caption}
\DeclareCaptionFormat{listing}{#1#2#3}
\mlsystitlerunning{PyTorch-Direct: Enabling GPU Centric Data Access for Very Large Graph Neural Network Training with Irregular Accesses}

\begin{document}

\twocolumn[
\mlsystitle{PyTorch-Direct: Enabling GPU Centric Data Access for Very Large Graph Neural Network Training with Irregular Accesses}




\begin{mlsysauthorlist}
\mlsysauthor{Seung Won Min}{uiuc}
\mlsysauthor{Kun Wu}{uiuc}
\mlsysauthor{Sitao Huang}{uiuc}
\mlsysauthor{Mert Hidayetoğlu}{uiuc}
\mlsysauthor{Jinjun Xiong}{ibm}
\mlsysauthor{Eiman Ebrahimi}{nvidia}
\mlsysauthor{Deming Chen}{uiuc}
\mlsysauthor{Wen-mei Hwu}{uiuc}
\end{mlsysauthorlist}

\mlsysaffiliation{uiuc}{University of Illinois at Urbana-Champaign, IL, USA}
\mlsysaffiliation{ibm}{IBM T.J. Watson Research Center, NY, Yorktown Heights}
\mlsysaffiliation{nvidia}{NVIDIA, TX, Austin}

\mlsyscorrespondingauthor{Seung Won Min}{min16@illinois.edu}

\mlsyskeywords{Machine Learning, MLSys}

\vskip 0.3in

\begin{abstract}
With the increasing adoption of graph neural networks (GNNs) in the machine learning community, GPUs have become an essential tool to accelerate GNN training.
However, training GNNs on very large graphs that do not fit in GPU memory is still a challenging task.
Unlike conventional neural networks, mini-batching input samples
%
in GNNs requires complicated tasks such as traversing neighboring nodes and gathering their feature values. 
While this process accounts for a significant portion of the training time, we find existing GNN implementations using popular deep neural network (DNN) libraries such as PyTorch are limited to a CPU-centric approach for the entire data preparation step.
This ``all-in-CPU'' approach has negative impact on the overall GNN training performance as it over-utilizes CPU resources and hinders GPU acceleration of GNN training.
To overcome such limitations, we introduce PyTorch-Direct, which enables a GPU-centric data accessing paradigm for GNN training.
%
In PyTorch-Direct, GPUs are capable of efficiently accessing complicated data structures in host memory directly without CPU intervention.
%
Our microbenchmark and end-to-end GNN training results show that PyTorch-Direct reduces data transfer time by 47.1\% on average and speeds up GNN training by up to 1.6$\times$.
Furthermore, by reducing CPU utilization, PyTorch-Direct also saves system power by 12.4\% to 17.5\% during training.
%
To minimize programmer effort, we introduce a new ``unified tensor'' type along with necessary 
changes to the PyTorch memory allocator, dispatch logic, and placement rules. 
As a result, users need to change at most two lines of their PyTorch GNN training code for each tensor object to take advantage of PyTorch-Direct.
%

\end{abstract}
]



\printAffiliationsAndNotice{} 

\section{Introduction}
\label{sec.intro}

Graphs are widely used to represent relational information between different entities.
Many real-world applications such as social networks, e-commerce, and logistics networks use graphs to track the relationship between human to human, human to product, location to location, and so on.
To take full advantage of this relational information, there has been increasing interest in adopting graph neural networks (GNNs) to enable various kinds of analysis such as node classification, link prediction, and graph property prediction~\cite{hu2020ogb} on graph-based databases.

\begin{figure}[t]
  \centering
  \includegraphics[width=\linewidth]{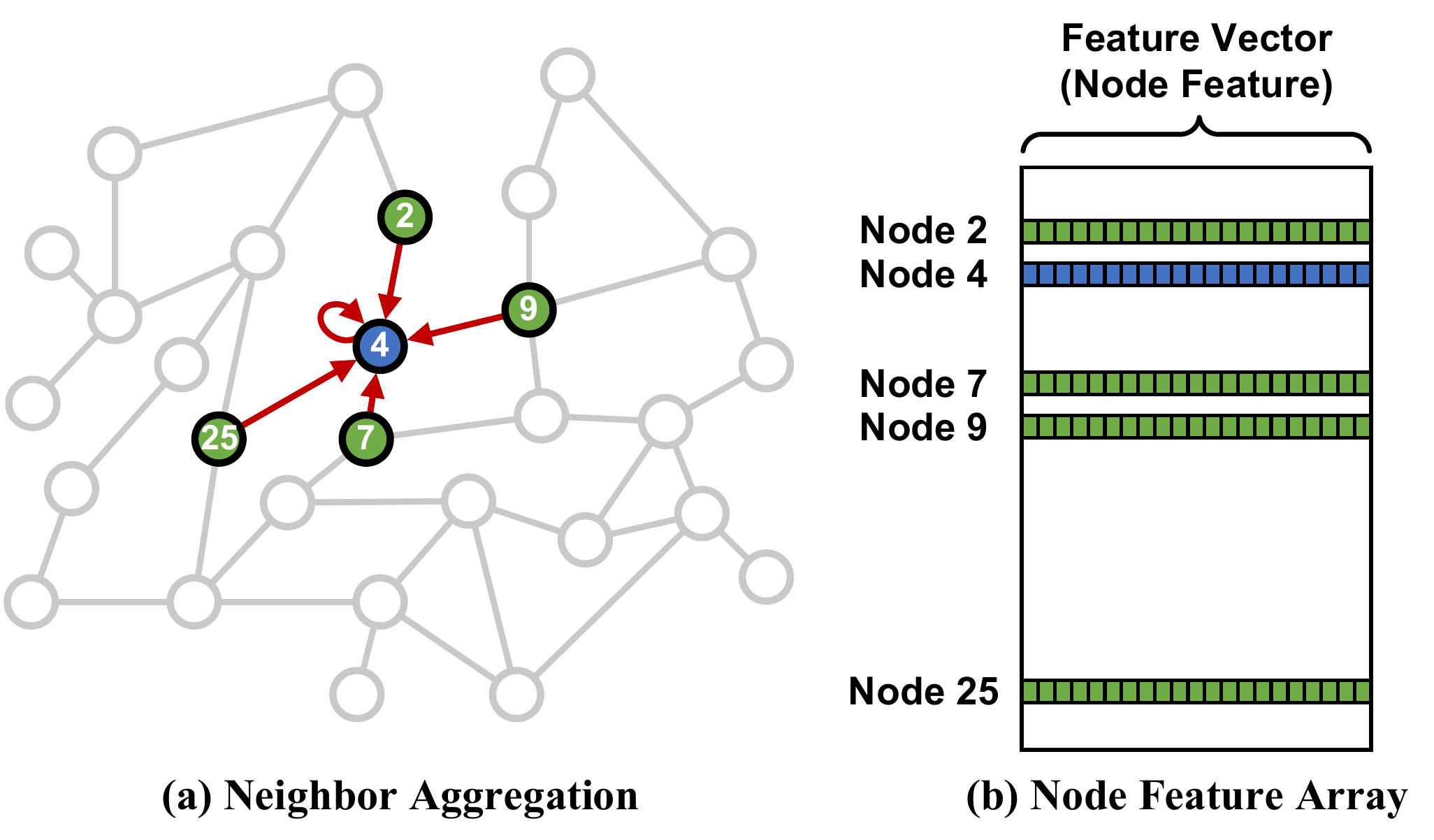}
  \caption{(a) A simple example showing feature aggregation in GNN training. (b) An example of node features in memory. The nodes might be close to each other in the graph, their features are not necessarily in memory.}
  \label{fig:background}
  \vspace{-4mm}
\end{figure}

In Figure~\ref{fig:background} (a), we show a simple example of GNN training.
To generate the embedding of node 4,
we traverse the input graph and gather information (features) of neighboring nodes 2, 4, 7, 9, and 25.
The gathered features go through a GNN architecture dependent aggregator to generate a state.
In this case, we assume node 4 is a root node, and the state generated by the first layer of aggregation is the final state, which is also a node embedding.
%
This process can be done multiple times recursively to look up further nodes.

%

With the advent of GNNs, there have been several efforts to provide generalized GNN training frameworks~\cite{StellarGraph,grattarola2020graph,wang2019dgl,Fey/Lenssen/2019} in popular Python-based DNN libraries such as PyTorch~\cite{adam2019pytorch} and TensorFlow~\cite{tensorflow2015-whitepaper}.
By using already existing Python-based DNN libraries, the new GNN frameworks are able to provide a new learning environment to the general user without a steep learning curve.

However, directly applying the DNN libraries for GNN training also introduces a new kind of performance challenge due to their CPU-centric way of transferring data between CPU and GPU.
%
Unlike training classical neural networks such as convolutional neural networks (CNNs), a single forward propagation of GNNs requires reading hundreds or even thousands of separate samples (node features) depending on the input graph topology
and the number of GNN layers.
%
Accessing the features of neighboring nodes in graphs is not as simple, i.e. not as structured,
as accessing the neighboring pixel values in images. 
Due to the way graph structures are stored in memory, features that the training algorithm accesses in order to process the output for a given node are likely scattered in memory, as depicted in Figure~\ref{fig:background}(b).

%

Now, the main challenge arises when we attempt to accelerate GNN training by using GPUs.
Similar to training other DNNs, GNN training also requires a huge number of arithmetic operations while aggregating the features, which makes the usage of GPUs attractive. 
%
Unfortunately, transferring the scattered data to the GPUs with the existing DNN libraries is not straightforward.
Initiating a direct memory access (DMA) call on each data fragment is too expensive and therefore the scattered data needs to be first gathered by the CPUs before the transfer.
For small graphs, this inefficiency can be bypassed by simply loading the whole features into GPU memory, but real world graphs can go beyond billions of nodes~\cite{ying2019pinsage} and thus far exceed GPU memory capacity.

In GNN training, the gather operation is not the only workload for CPUs.
CPUs need to generate subgraphs for each mini-batch and constantly traverse input graphs to identify neighboring nodes.
%
The number of these graph structure related operations and the portion of total training time they consume is non-trivial (44\%-99\%)~\cite{liu2020g3}.
%
While CPU resources are already overloaded, it is impractical to further burden CPUs with data transfer operations.
%

Conventional wisdom would argue that since the graph feature data is in host memory, the CPU should have significant latency and bandwidth advantage over GPUs in performing the gather operations on these features. However, GPUs with their ability to issue a massive number of concurrent memory accesses to tolerate latency, have been recently shown to be effective in accessing data with irregular structures like graphs that reside in the host memory~\cite{min2020emogi}. If successful, having the GPUs perform gather operations also eliminates the need to perform a data copy from the CPU to the GPU after the feature data has been gathered.
%
It is therefore desirable to explore the use of GPUs to perform feature gather accesses to significantly reduce end-to-end GNN training time. 
In this work, we present PyTorch-Direct, a GPU-centric data access design for GNN training.

PyTorch-Direct presents a new class of tensor called ``unified tensor.'' 
While a unified tensor resides in host memory, its elements can be accessed directly by the GPUs, as if they reside in GPU memory.  
To support seamless transition of applications from the original PyTorch to PyTorch-Direct, we design the programming interface for using unified tensors to be consistent with the existing PyTorch GPU tensor declaration mechanism.
Users can take full advantage of PyTorch-Direct, by merely modifying 2 lines of their PyTorch GNN implementation.
%

PyTorch-Direct also includes an important memory access alignment optimization for direct host memory access over PCIe.
Without the aligned memory accesses, direct access over PCIe could suffer performance drop of nearly 44\%.
%
To prevent this, we implement a circular shift stage in PyTorch GPU indexing function which automatically calculates the offset values required to enforce alignment and generate adjusted indices.
The adjustment algorithm is entirely implemented in the PyTorch backend CUDA code and general users do not need to do anything special in their
code to take advantage of it.

With PyTorch-Direct, the time spent for accessing irregular data structures in host memory is reduced on average by 47.1\% compared to the baseline PyTorch approach.
In real GNN training, we show PyTorch-Direct can speedup end-to-end GNN training by up to 1.62$\times$ depending on GNN architecture and input graph.
Furthermore, by reducing the CPU workload, PyTorch-Direct provides 12.4\% to 17.5\% of reduced system power consumption during GNN training.


%
We designed PyTorch-Direct targeting high performance, portability, and ease of use. 
%
%
The main contributions of this paper are summarized as follows:
\begin{itemize}
\vspace{-1mm}
    \item We identify inefficient host to GPU data transfer patterns in existing GNN training schemes that cause high CPU utilization and increase end-to-end training time.
\vspace{-1mm}
    \item We propose
    a GPU-centric data access paradigm with a novel circular shift indexing optimization for GNN training to reduce training time, CPU utilization, and power consumption.
\vspace{-1mm}
    \item We incorporate the proposed system level changes seamlessly into a popular DNN library, PyTorch, with a comprehensive implementation to benefit a broad range of GNN architectures.
\end{itemize}
\section{Background and Related Work}
\label{sec.related}

\subsection{Graph Neural Networks}
\label{sec.background.gnn}
Graph neural network (GNN) is a type of neural network that works on graph data.
Unlike the conventional neural networks, GNNs can retain the relational information between entities represented by the graph nodes.
A common example of GNN application is a recommender system in social networks~\cite{hamilton2017inductive,ying2019pinsage,kipf2017semi}.
GNNs are largely composed of two types of components: aggregators and updaters~\cite{zhou2019gnn}.
Aggregators gather features from each node's neighbors and updaters update nodes' hidden states.
%
Together, the process of outputting a state embedding $h_v$ of an arbitrary node $v$ can be expressed as follows:


\vspace{-4mm}
\begin{equation}
\label{eq.gnn_formulat}
h_v = f(x_v, x_{co[v]}, x_{ne[v]}, h_{ne[v]})
\vspace{-2mm}
\end{equation}



Here, $f$ is a parametric function.
$x_v$, $x_{co[v]}$, and $x_{ne[v]}$ are the input features of the node $v$, edges connected to the node $v$, and neighboring nodes of node $v$, respectively, whereas
$h_{ne[v]}$ represents
the states of neighboring nodes of node $v$.
%
This process can be done recursively in a bottom-up tree traversal manner to aggregate features and states of nodes several hops away.
%
The final output embeddings can be used for many downstream tasks, such as node classification.

%
%


%
%
%
%
%

\subsection{GPU Out-of-memory Solution for GNN Training}
In GNN training, the input features are located in a 2D array where the row indices are the IDs of nodes and the columns are the features of each node.
%
In Figure~\ref{fig:background}, we show a case of retrieving the node features of the neighboring nodes during the GNN training.
Due to the structural discrepancy between the graph and the array, accessing the features of neighboring nodes in the graph results in accessing rather unpredictable and non-sequential rows of the Feature Array.

A straightforward approach to sending these non-consecutive rows to the GPU is to call data copying functions like \texttt{cudaMemcpy()} multiple times, once for each row. 
Unfortunately, 
making multiple calls to data copying functions incurs significant overhead and can be highly inefficient. 
When the input graphs are small, one can bypass this issue by simply placing
the entire feature array into the GPU memory at the beginning of GNN training.
However, in reality, it is not reasonable to assume the the entire feature array can always fit into the GPU memory.

Currently, the solutions for training GNNs on very large graphs can be largely divided into two categories:
1) Only the immediately necessary features for the current mini-batch are gathered by the CPU, and then sent to the GPU memory~\cite{ying2019pinsage}.
%
2) Before training, partition the input graphs into multiple smaller subgraphs that can be fit into the GPU memory, and then train on them one by one~\cite{chiang2019cluster,graphsaint-iclr20}.
In the former category, the CPU can become a bottleneck and slows down the training pipeline. In the latter category, the subgraphs inevitably lose some of the distinct structural patterns of the original graphs~\cite{zu2019survey}.
Pytorch-Direct addresses these deficiencies by enabling the GPU to directly gather all the needed features from the host memory on demand.

\subsection{GNN Frameworks with Python DNN Libraries}
To simplify GNN development, there are several efforts to combine some of the commonly required functionalities in GNN trainings and create a framework using popular Python-based DNN libraries such as PyTorch and TensorFlow.
Deep graph library (DGL)~\cite{wang2019dgl} is developed based on MXNet, PyTorch, and TensorFlow. 
PyTorch-Geometric~\cite{Fey/Lenssen/2019} is a PyTorch-based GNN framework.
StellarGraph~\cite{StellarGraph} and Spektral~\cite{grattarola2020graph} are based on TensorFlow and Keras API. In this work, we demonstrate the benefit of our approach by extending PyTorch.
\section{Motivation}
\label{sec.motivation}

\begin{figure}[t]
  \centering
  \includegraphics[width=0.85\linewidth]{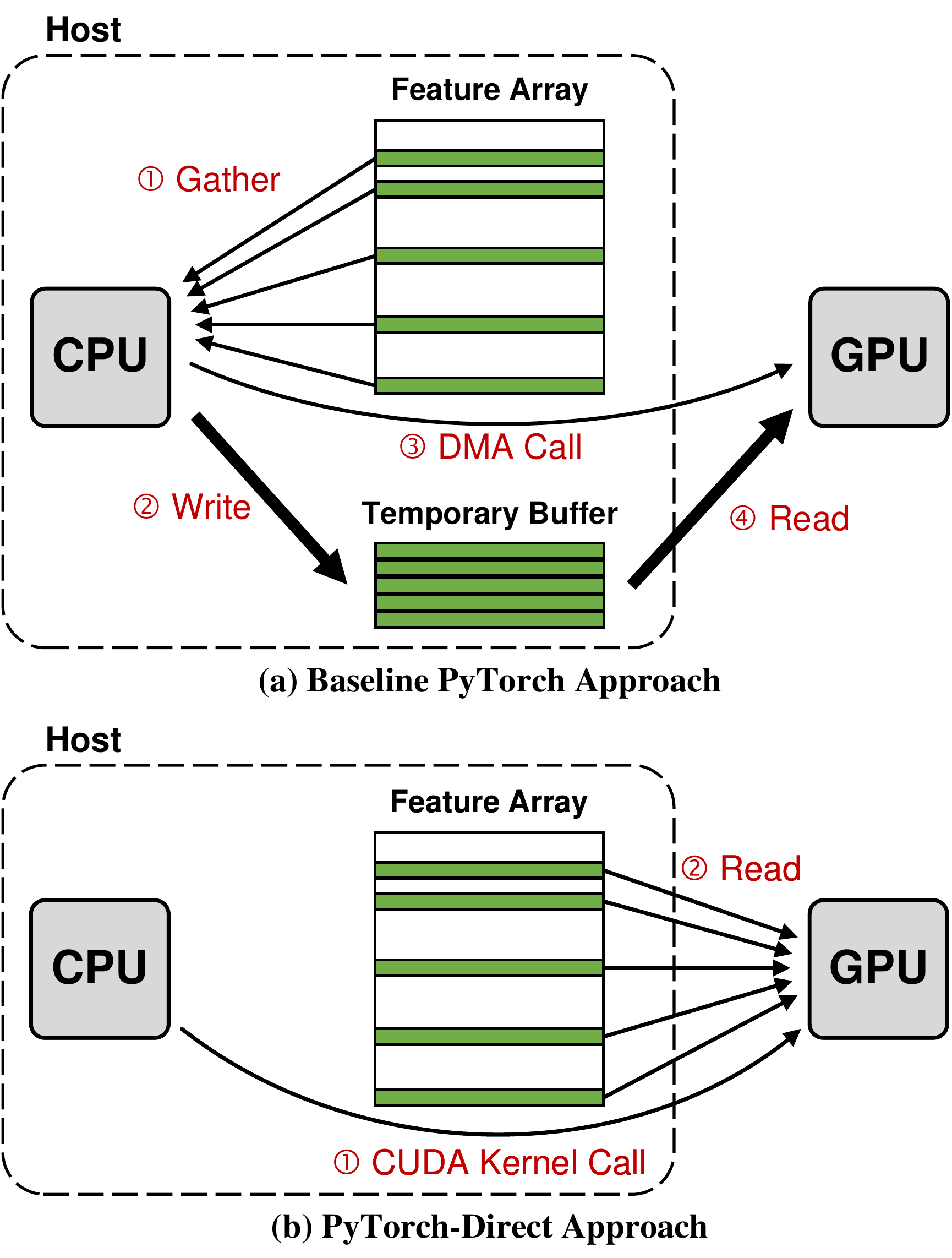}
  \caption{(a) High level depiction of data transfer mechanism in current PyTorch implementation. (b) Simplified data transfer mechanism in PyTorch-Direct with direct access.}
  \label{fig:design_comparison}
  \vspace{-4mm}
\end{figure}    

%

%
In current implementations of deep learning frameworks, host to GPU data loading process is implemented in a CPU-centric manner.
%
When data that needs to be processed by the GPU is scattered in host memory, it is the CPU's responsibility to gather the data fragments before calling a DMA.  Figure~\ref{fig:design_comparison} (a) shows the four main steps of this CPU-centric approach. The CPU first reads (gathers) the features, i.e. relevant rows of the Feature Array in this example, into its cache (\circled{1}), 
it then writes them into consecutive locations in a temporary buffer (\circled{2}) 
before it calls a data copy function to set up a DMA operation (\circled{3}) 
and finally, the DMA hardware on the GPU reads the data from the temporary buffer in host memory into a corresponding buffer in the GPU memory (\circled{4}). 
%

\begin{figure}[t]
  \centering
  \includegraphics[width=\linewidth]{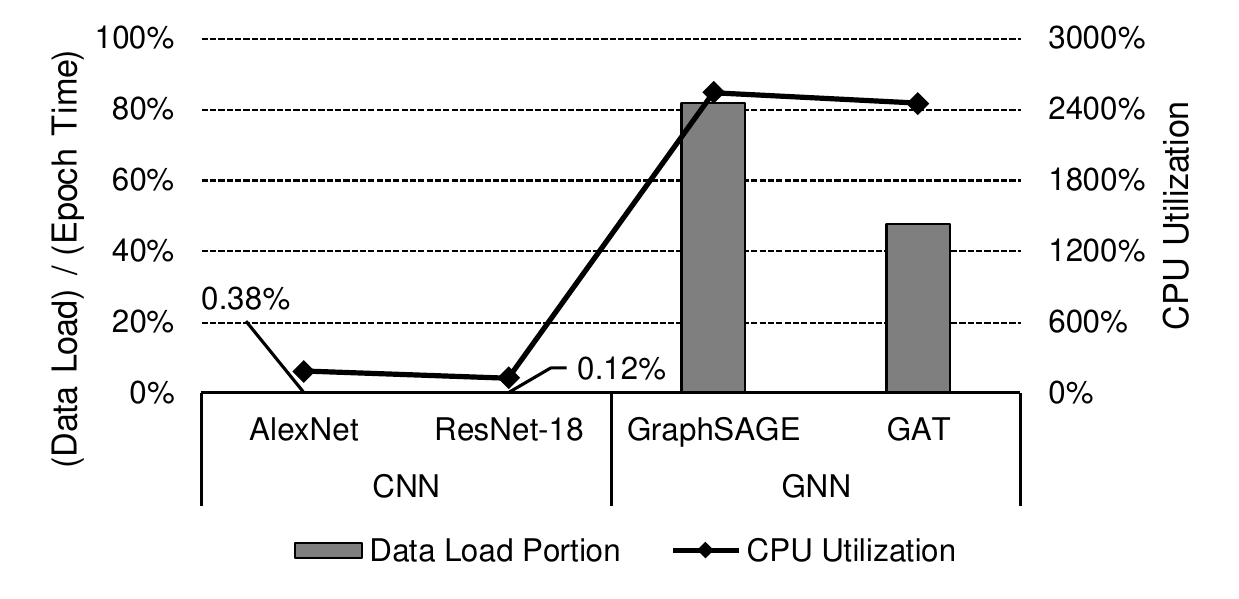}
  \caption{CPU utilization and data loader time comparison between CNN and GNN trainings. CPU utilization can go beyond 100\% as it is multithreaded.}
  \label{fig:motivation}
  \vspace{-4mm}
\end{figure}

In Figure~\ref{fig:motivation}, we show the impact of this CPU-centric data loading approach on GNN training.
As a comparison, we use AlexNet~\cite{krizhevsky2012imagenet} and ResNet-18~\cite{He2015} as CNN examples and GraphSAGE~\cite{hamilton2017inductive} and graph attention network (GAT)~\cite{attention2018graph} as GNN examples.
We use Torchvision~\cite{torchvision} for CNN training and DGL backed by PyTorch for GNN training.
While the time spent for data loading is less than 1\% of the CNN training time, it consumes 47\% and 82\% of the GNN
training time for GrapSAGE and GAT, respectively.
As the vertical axis on the right of Figure~\ref{fig:motivation} shows, CPU utilization is also much higher in GNN training.
This happens partly because the data gathering part of the code is multithreaded and tries to maximize the throughput and thus minimize latency.
%
Additionally, multi-threading is also used to maximize 
the performance of graph traversal and subgraph generation during data loading.
%

In short, in GNN training, unlike CNN training, data loading incurs significant time and resource overheads. In this work, we aim to reduce this overhead that comes from inefficient use of CPU resources in gather operations.
%
%
We propose a GPU-centric approach to accessing data for GNN training based on the direct host-memory-access capability of modern GPUs (Figure~\ref{fig:design_comparison} (b)).
Modern GPUs have their own address translation units and are capable of accessing host memory directly.
If GPUs are connected over PCIe, they can simply generate PCIe read/write I/O
requests to the host.
From the programmer's point of view, accessing host memory can be simply done by dereferencing unified memory pointers just like dereferencing device memory pointers.

This direct access feature is different from the conventional
unified virtual memory (UVM) method which is based on page migration.
In UVM, the data transfer between host and GPU is done in page granularity which is at least 4kB per page in modern computing systems.
%
Whenever a required page is missing from the GPU, the CPU needs to handle the page fault through a hardware interrupt service.
Since the minimum data transfer granularity is a page
and the hardware interrupt service process is costly, 
the performance of the UVM method depends on the applications' spatial and temporal localities~\cite{uvmguide}.
When dealing with highly irregular data structures such as a graph, using UVM incurs excessive page-faults and I/O amplification~\cite{Gera20,min2020emogi,Sabet20}.

In the following section, we describe our implementation of PyTorch-Direct which enables GPU-centric data accesses for the PyTorch DNN library.
We mainly focus on PyTorch in this work due to its straightforward and intuitive way of binding data to a certain physical location from the user's perspective.
But the main idea of the GPU-centric data accessing mechanism can still be applied to the other DNN frameworks such as TensorFlow.

%
%
%
%
%

\section{PyTorch-Direct}
\label{sec.PyTorch_direct}

This section describes the design and implementation of PyTorch-Direct.
First, we provide an overview of design goals and introduce a new type of tensor, i.e., \textit{the unified tensor}, which incorporates new concepts in need.
We then discuss the unified tensor API and its advanced configurations.
Finally, we describe our implementation and optimizations.

\subsection{Overview}
\label{sec.PyTorch_direct.Overview}


PyTorch-Direct aims to enable GPU out-of-memory training and inference for GNN while incorporating the direct access feature to improve data access performance. 
To achieve this, PyTorch-Direct presents to the developers several API features that are centered around a new type of tensor called ``unified tensor''.
It is a new, independent type parallel to PyTorch native GPU or CPU tensors, from both the perspective of user interface and its implementation in the runtime system. We have developed all the supporting code that allows unified tensors to be used as a full-fledged type of tensor in all PyTorch runtime activities such as memory allocator, \texttt{torch.device} class, dispatch, etc. This makes it extremely easy for the application developers to adapt their PyTorch code to use unified tensors.
%

%
%

Unified tensors are at the core of the PyTorch-Direct design which enables GPUs to directly operate on the host memory.
%
Both of the underlying CUDA and CPU C++ codes of PyTorch runtime code can directly access
unified tensors by simply dereferencing unified memory pointers.
In comparison, PyTorch native CPU tensors can be only accessed by CPU and GPU tensors can be only accessed by GPU, thus limiting the type of computation devices that can participate in the processing of these tensors. 
Unified tensors eliminate these limitations.

By default, PyTorch-Direct physically allocates the unified tensors in the host memory and allows GPUs to directly access them over the PCIe.
Since the unified tensors are located in the host memory, their sizes can grow beyond the GPU memory size.
From the CPU's perspective, accessing the unified tensors is identical to accessing CPU tensors. 
%
%
%
%

\begin{lstlisting}[label=unified-example-original,caption=An example of GNN training in PyTorch.,frame=tb,float=]
 # Load features into regular CPU tensor
 features = dataload()
  
 for epoch in range(num_epochs):
   for (neighbor_id, |\ldots|) \
     in enumerate(neighbor_sampler):
      
     # Gather features using neighbor_id
     # and then copy to GPU
     input_features = \
       features[neighbor_id].to("cuda")
          
     train(input_features, |\ldots|)|\lstsetnumber{\ldots}|
     |\ldots||\lstsetnumber{}||\lstresetnumber\setcounter{lstnumber}{299}|

\end{lstlisting}

%
%
%
Application developers can adapt their PyTorch code to use unified tensors with minimal change to their code.
In Listing \ref{unified-example-original}, we show a simplified example of GNN training in PyTorch.
%
After loading all the features into host memory, in every training step, it sends the features in the minibatch to GPU by calling \texttt{to("cuda")} before invoking the \texttt{train} function (line 10--13).

The procedure with the unified scheme is shown in Listing \ref{unified-example}.
In this example, to migrate to the unified tensor scheme, the developer only needs to remove the \texttt{to("cuda")} invocation on \texttt{features[neighbor\_id]} and instead invokes \texttt{to("unified")} on \texttt{features} at the beginning.
The features of the whole graph is now stored in a unified tensor that can hold data beyond the GPU memory capacity.
After that, GPU kernels that are launched by the \texttt{train} function can directly access \texttt{features} since it can access unified tensor and its derived tensors.
Therefore, no more \texttt{to()} invocation is needed.
Section \ref{sec.PyTorch_direct.API} describes more about the API design, including advanced configurations.
%

%

The fact that unified tensor is a full-fledged type in PyTorch-Direct enables clean implementation of complicated rules in runtime systems and easy future extensions.
For example, PyTorch-Direct clearly defines the whole set of rules to resolve computation placement and output tensors placement for computation that involves unified tensors, as detailed in Section \ref{subsec.PyTorch_direct.place_rule}. Thanks to the completeness of unified tensor, this is well-integrated into the PyTorch runtime system. The implementation details are discussed in Section \ref{sec.PyTorch_direct.implementation},

\begin{lstlisting}[label=unified-example,caption=GNN training in PyTorch-Direct with unified tensor. Only two lines (2 and 11) from Listing \ref{unified-example-original} are changed to incorporate unified tensor .,frame=tb,float=]
 # Load features into unified tensor
 features = dataload().to("unified")
  
 for epoch in range(num_epochs):
   for (neighbor_id, |\ldots|) \
     in enumerate(neighbor_sampler):
      
     # GPU directly fetches required
     # features from unified tensor
     input_features = \
       features[neighbor_id]
          
     train(input_features, |\ldots|)|\lstsetnumber{\ldots}|
     |\ldots||\lstsetnumber{}||\lstresetnumber\setcounter{lstnumber}{299}|
\end{lstlisting}

\subsection{API Design}
\label{sec.PyTorch_direct.API}

\begin{table*}[t]
\caption{Typical usage of APIs with unified tensor. Unified tensors are allowed for easy creation and flexible computation.}
\label{tab:APIUnified}
\vskip 0.15in
\begin{center}
\begin{small}
\begin{tabular}{ll}
\toprule
Example & \multicolumn{1}{c}{Description}\\
\midrule
\texttt{one\_tensor.to("unified")} & Copy the tensor \texttt{one\_tensor} to unified device.\\
\texttt{torch.ones(128, device="unified")} & Specify unified device in PyTorch native APIs.\\
\texttt{one\_tensor.is\_unified} & Returns \texttt{True} if the variable \texttt{one\_tensor} is a unified tensor.\\
\texttt{unified\_tensor + cpu\_tensor} & Computation with hybrid tensors of unified and CPU types.\\
\texttt{unified\_tensor[gpu\_tensor]} & Indexing unified tensor with GPU tensor.\\
\bottomrule
\end{tabular}
\end{small}
\end{center}
\vskip -0.1in
\end{table*}

PyTorch-Direct APIs are designed to provide interface to unified tensors in the idiomatic PyTorch manner.
%
%
Table \ref{tab:APIUnified} demonstrates typical use of unified tensor APIs.
Developers can create a unified tensor by copying from another tensor via PyTorch built-in \texttt{to()} method of \texttt{torch.Tensor}.
It can also be created from scratch by specifying the \texttt{device} argument as unified device in PyTorch APIs, such as \texttt{torch.ones}.
The user can check if a tensor is of unified type by invoking the \texttt{is\_unified} method.

Unified tensors can be computed with CPU or GPU tensors, providing great flexibility.
%
%
Meanwhile, they are free from redundant data movements since CPU and GPU can directly access their underlying memory without creating temporary copies.
By contrast, in the native PyTorch API, CPU tensors typically cannot work with GPU tensors because of the device binding, unless additional routines to handle them have been implemented manually in PyTorch runtime system.
%
For example, subscript operator allows a GPU tensor to be indexed by a CPU tensor, and binary and comparison operators accept GPU scalar and CPU scalar as the two operands.
%

PyTorch-Direct also exposes two configurations, placement rule hints and \texttt{cudaMemAdvise} hints, so that experienced developers can perform advanced optimizations. 
%
These hints can be used in a handful of APIs, as shown in Table \ref{tab:APIUnifiedAdvanced}.

\begin{table*}[t]
  \vspace{1mm}
  \caption{Advanced unified tensor APIs with default arguments values. They set up \texttt{cudaMemAdvise} and placement hints as detailed in Section \ref{subsec.PyTorch_direct.place_rule}. New keywords and methods are in bold.}\vspace{-2mm}
  \label{tab:APIUnifiedAdvanced}
  \vskip 0.15in
  \begin{center}
  \begin{small}
  \begin{tabular}{rll}
  \toprule
  \multicolumn{2}{l}{Advanced API} & \multicolumn{1}{c}{Description}\\
  \midrule
  \multicolumn{2}{l}{\texttt{torch.device("unified", \textbf{propagatedToCUDA=True})}} & Yield \texttt{torch.device} with placement hint flag set up.\\
  \multicolumn{2}{l}{\texttt{unified\_tensor.\textbf{set\_propagatedToCUDA}(bool\_val)}} & Switch the placement hint flag of unified tensor.\\
  \texttt{one\_tensor.to(}&\texttt{unified\_device,} & \multirow{3}{7cm}{Create unified tensor with \texttt{cudaMemAdvise} hint invoked onto its storage.}\\
  &\texttt{\textbf{advise='SetPreferredLocation'},}& \\
  &\texttt{\textbf{adviseDevice='cpu'})} &  \\
  \multicolumn{2}{l}{\texttt{unified\_tensor.\textbf{memAdvise}(advise, adviseDevice)}} & Invoke \texttt{cudaMemAdvise} and returns error code.\\
  \bottomrule
  \end{tabular}
  \end{small}
  \end{center}
  \vskip -0.1in
  \end{table*}

One configuration is the placement rule hint flag, \texttt{propagatedToCUDA}, that is owned by each unified tensor. The placement rules will be detailed in Section \ref{subsec.PyTorch_direct.place_rule}.
%
%
%
%
%
%
PyTorch-Direct allows the developer to specify the placement rule hint during creation, and switch in the middle.
%
%
 %
 Switching is provided by the \texttt{set\_propagatedToCUDA} method.
 It only switches the placement rule of a unified tensor without memory allocation, deallocation or data copy.
 %
 Though this is a \texttt{torch.Tensor} method,  invoking this method on a non-unified tensor triggers \texttt{RuntimeError}.
 %
 %

As for \texttt{cudaMemAdvise}, PyTorch provides the interface to suggest the CUDA runtime to populate, migrate or prefetch data on the specified device.
This can be done by invoking the \texttt{memAdvise} method of \texttt{torch.Tensor}.  
The two parameters of the \texttt{memAdvise} method are the same as \texttt{cudaMemAdvise}, though they are in the form of Python \texttt{str} and \texttt{torch.device}.
PyTorch-Direct runtime converts the arguments and invokes \texttt{cudaMemAdvise}.
Similar to \texttt{set\_propagatedToCUDA}, the \texttt{memAdvise} method triggers a \texttt{RuntimeError} if the argument tensor is not of unified type.

Users may also specify the \texttt{cudaMemAdvise} hint as optional arguments in tensor creation methods invocation, just like in the case of placement rules. In such cases, PyTorch-Direct invokes \texttt{cudaMemAdvise} right after the memory allocation. On the other hand, the developer may instead invokes the tensor's \texttt{memAdvise} method after its creation. However, the implementation of some PyTorch tensor creation APIs involves GPU accesses to the new tensor before returning to the Python interpreter, bringing in side effects to the CUDA runtime. In such cases, calling \texttt{memAdvise} method instead of specifying \texttt{cudaMemAdvise} keywords in creation cannot properly affect the GPU accesses that happen before the creation function returns, which can lead to unexpected \texttt{cudaMemAdvise} outcomes.
%

%
%


\subsection{Computation and Storage Placement Rules}
\label{subsec.PyTorch_direct.place_rule}


Although the unified tensor type abstracts away the binding to physical devices, the computation of each operator needs to be executed in one of the physical CPU or GPU devices.
The runtime also needs to determine which one of the physical devices should hold the output tensors.
Moreover, in cases where the output tensors are to be reused for multiple times by a GPU device, it will be more performant to explicitly transfer them to the GPU device's memory, instead of having the GPU repetitively perform direct access to them from the host memory. 

Thus, it is important to clearly specify the computation and storage placement rules applied to each operation as well as to allow application developers to take advantage of data reuse opportunities. 
To achieve this, each unified tensor indicates its placement preference between two options with its \texttt{propagatedToCUDA} flag. The placement rule for each operator is determined dynamically. 

%

A unified tensor with \texttt{propagatedToCUDA} set to \texttt{True} prefers GPU as the computation device.
Besides, it prefers the output tensor type to be GPU or unified.
Placing computation to GPU and creating GPU output tensors give the best performance when the output tensor is reused after the computation. 
A unified tensor with \texttt{propagatedToCUDA} set to \texttt{False} prefers the opposite.

%
%

%

%

%
%

\begin{table*}[t]
\caption{Placement Rules Applied to Operators with Unified Tensors Operands}
\label{tab:place_rule}
\vspace{-2.5mm}
\begin{center}
\begin{small}
\resizebox{\linewidth}{!}{
\begin{tabular}{llll}
\toprule
& \multicolumn{2}{l}{All unified tensors prefer propagation.}  & At least one unified tensor prefer non-propagation.       \\
\midrule
At least one operand is non-scalar   & \multicolumn{1}{l|}{compute on} & GPU  & CPU if no operand prefers propagation, else GPU\\
CPU tensor. & \multicolumn{1}{l|}{output type}& unified non-propagation & unified non-propagation \\
\hline
Previous row is not applicable. And,   & \multicolumn{1}{l|}{compute on} & GPU  & GPU \\
 at least one is GPU tensor operand. & \multicolumn{1}{l|}{output type}& GPU   & unified propagation \\
\hline
\multirow{2}{5cm}{All non-unified tensors are CPU scalars. Or, no non-unified tensors exist.}   & \multicolumn{1}{l|}{compute on} & GPU  & CPU if no operand prefers propagation, else GPU \\
 & \multicolumn{1}{l|}{output type} & GPU   & unified non-propagation \\
\bottomrule
\end{tabular}
}
\end{small}
\end{center}
\vspace{-4mm}
\end{table*}
%
Complications arise when 
the operands of an operation are of different tensor types or indicate opposite preferences.
%
%
The complete placement rules are shown in Table \ref{tab:place_rule}.
It uses ``propagation'' to refer to unified tensor with \texttt{propagatedToCUDA} set to \texttt{True}, and ``non-propagation'' to indicate otherwise.
The table lists the computation device and output tensors type in all the six scenarios, with rows specifying the condition of non-unified tensors and columns specifying the condition of unified tensors.



    %

%

\subsection{Implementation Details}
\label{sec.PyTorch_direct.implementation}


The core object in PyTorch runtime system is \texttt{at::Tensor}.
Every
PyTorch tensor (\texttt{torch.Tensor} object) is a \texttt{THPVariable} object in C++ runtime code, which is the wrapper class combining an \texttt{at::Tensor} object with Python metadata.
The PyTorch runtime dispatches each method call to the right definition according to the device and data types of the tensor arguments.
A PyTorch method operating on tensors goes eventually into a function of \texttt{at::Tensor}.
%
%

%
PyTorch-Direct implements the unified tensor mechanisms in PyTorch runtime as a complete type of tensor.
This allows the design to be modular, extensible, and well integrated into the PyTorch runtime code.
%
%
A new memory allocator is implemented to govern the memory allocation for all unified tensors.
It adapts the allocation recycling mechanism from the PyTorch CUDA allocator to reduce the number of CUDA API invocations.

Two dispatch keys are introduced to the runtime system. 
They each represent either state of the \texttt{propagatedToCUDA} flag of the unified tensor and participate in resolving the placement rules of operators, as detailed in Section \ref{subsec.PyTorch_direct.place_rule}.
Therefore,
PyTorch-Direct now dispatches invocations with unified tensors according to the dispatch table of unified tensors, allowing future extensions.
PyTorch-Direct in most cases dispatches to existing CPU or CUDA definitions because they can directly access the memory underlying unified tensors without modifications.
Nevertheless, we augment the logic of a few tensor creation methods to use unified memory allocator in the cases where unified tensors should be created.
%
%

PyTorch-Direct also implements other facilities, such as \texttt{c10::DeviceType}s for unified device and auxiliary methods to determine if an \texttt{at::Tensor} object is of unified type.
They are helpful in developing new logic and reusing existing code.
For example, an existing GPU or CPU operator definition checks if tensors are of GPU or CPU type at the beginning.
%
Modifying such checks to allow unified tensor enables the same function definition to also work for unified tensors.

\subsection{Memory Alignment Optimization}
\label{sec.PyTorch_direct.implementation.alignment}

\begin{figure}[t]
  \centering
  \includegraphics[width=\linewidth]{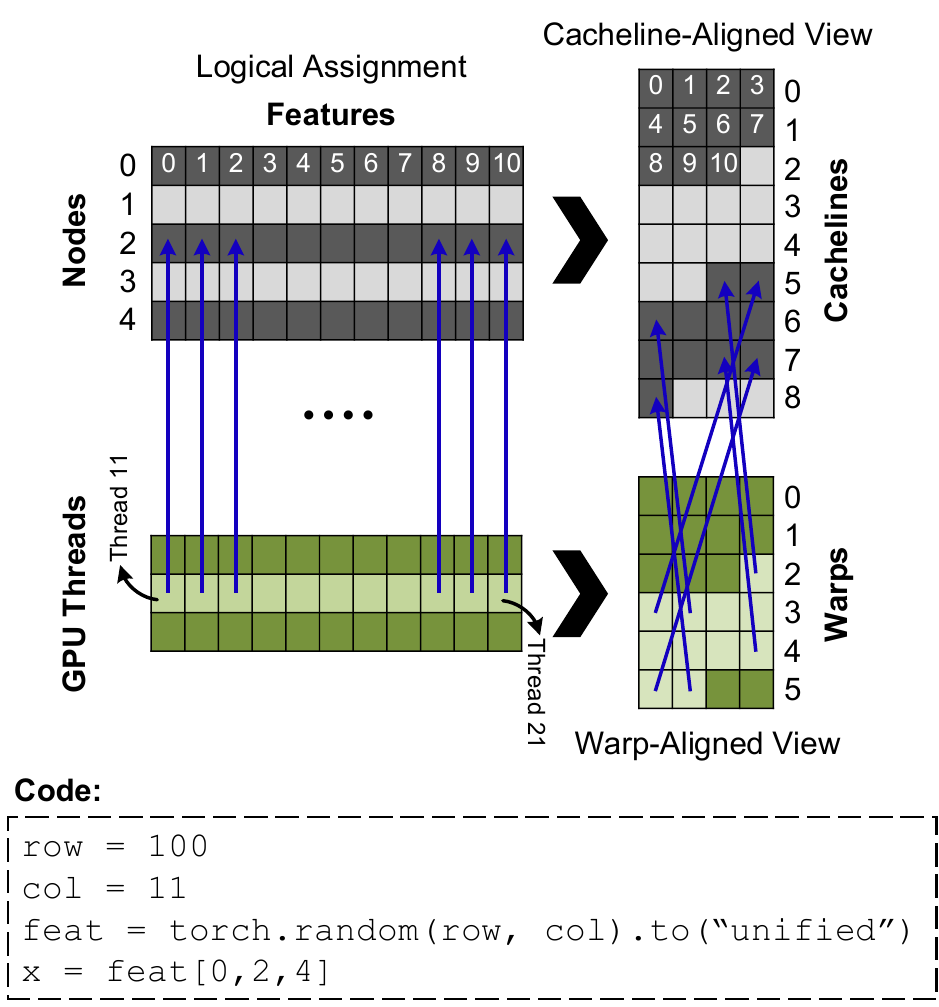}
  \caption{Data access misalignment occurring in PyTorch-Direct when using unmodified PyTorch indexing scheme. Based on the code, thread 0-10 access \texttt{feat[0]}, thread 11-21 access \texttt{feat[2]}, and thread 22-32 access \texttt{feat[4]}. For the case accessing \texttt{feat[2]} (blue arrows), we can easily identify the accesses are fragmented into multiple warps and cachelines.}
  \label{fig:alignment_problem}
  \vspace{0mm}
\end{figure}

To achieve efficient PCIe data transfer, memory requests from the GPU threads in the same warp should be aligned and merged to the GPU cacheline (128-byte) granularity~\cite{min2020emogi}.
%
However, the default PyTorch GPU indexing function does not guarantee the memory alignment unless the input feature tensors are naturally aligned with the GPU cacheline size.
In Figure~\ref{fig:alignment_problem}, we depict a simplified working mechanism of the default PyTorch GPU indexing function.
In this specific example, we scale down the warp size (32 threads in real) and the GPU cacheline size (128-byte in real) by a factor of 8. 
We assume each feature is 4-byte and each node has 11 features.
%
Now, due to the size mismatch between the cacheline (16-byte) and the node feature (44-byte), misaligned accesses can occur.
%

In the example of Figure~\ref{fig:alignment_problem}, assume the GPU needs to access the 0th, 2nd, and 4th nodes.
To achieve this, each thread accesses a single feature.
For example, the first 11 threads access the 11 features of the 0th node, next 11 threads access the 11 features of the 2nd node, and so on.
This looks simple in a logical view on the left side of Figure~\ref{fig:alignment_problem}, where we highlight the accesses of thread 11-21 to features of node 2. 
However, when we redraw the access patterns based on cacheline and warp alignments on the right side of Figure~\ref{fig:alignment_problem}, we can see that the accesses are fragmented into multiple cachelines and warps.

\begin{figure}[t]
  \centering
  \includegraphics[width=\linewidth]{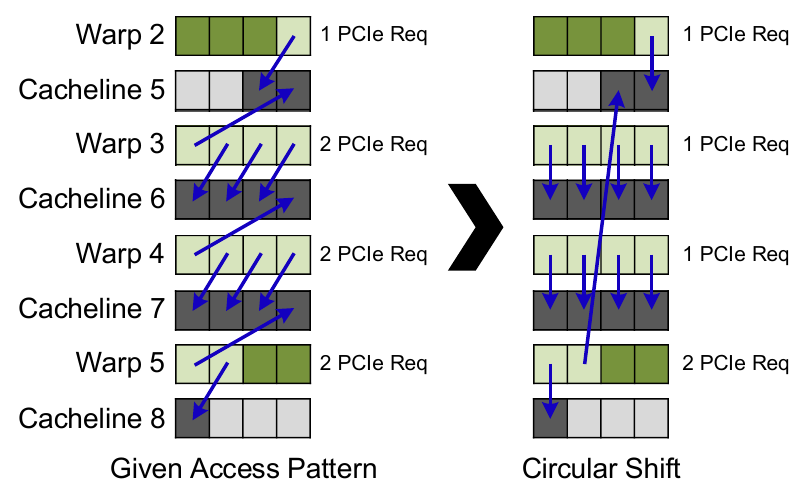}
  \caption{Memory alignment optimization with a circular shift. The example is identical to the case in Fig.~\ref{fig:alignment_problem}. Alignment reduces the total number of PCIe requests from 7 to 5 in this case.
  }
  \label{fig:alignment_fix}
\end{figure}

To solve the problem of misaligned access patterns, we use a circular shift method as described in Figure~\ref{fig:alignment_fix}.
In this method, all threads calculate the required index offset values to make aligned accesses.
In the case of Figure~\ref{fig:alignment_fix}, the threads need to do a right shift by an offset of 1.
The threads on the edges check the boundary conditions and make additional adjustments by adding or subtracting the length of the node feature so that they do not accidentally access the other node features.
When the indexed values are written to the output, the output indices are also identically adjusted to maintain the ordering.
With the optimization, PyTorch-Direct reduces the number of total PCIe requests from 7 to 5 in this case.
Inside the PyTorch GPU indexing kernel, we check the input tensors and apply this optimization only when the input tensors are unified tensors and the feature widths are not naturally aligned to 128-byte granularity.
All these adjustments are automatically made as a result of our modifications to PyTorch source code. As such, there is no programmer effort required for solving the memory alignment problem. 


\section{Evaluation}
\label{sec.evaluation}
In this section, we evaluate PyTorch-Direct performance using a well-defined microbenchmark and end-to-end GNN training.
Using the microbenchmark, we demonstrate that (1) PyTorch-Direct is faster than 
the baseline PyTorch approach
in accessing features from GPU under different combinations of data sizes and systems and (2) the effectiveness of our optimized memory alignment mechanism.
In GNN training, we show the benefit of using PyTorch-Direct for faster training with lower power consumption.

\subsection{Evaluation Setup}
\label{sec.evaluation.setup}
\setlength{\tabcolsep}{2pt}
\begin{table}[t]
\caption{Datasets for GNN training}
\label{tab:datasets}
\vspace{-2.5mm}
\begin{center}
\begin{tiny}
\resizebox{\linewidth}{!}{
\begin{tabular}{llcccc}
\toprule
Abbv.   & Dataset & \#Feat. & Size & \#Node & \#Edge \\
\midrule
reddit  & reddit                & 602 & 561MB    & 233.0K & 11.6M \\
product & ogbn-products         & 100 & 960MB    & 2.4M & 61.9M \\
twit    & twitter7              & 343 & 57GB    & 41.7M & 1.5B \\
sk      & sk-2005               & 293 & 59GB    & 50.6M & 1.9B \\
paper   & ogbn-papers100M       & 128 & 57GB    & 111.1M & 1.6B \\
wiki    & wikipedia\_link\_en   & 800 & 44GB    & 13.6M & 437.2M \\
\bottomrule
\end{tabular}}
\end{tiny}
\end{center}
\vspace{-2mm}
\end{table}

\textbf{Datasets:} The datasets we use for the GNN training evaluation are shown in Table~\ref{tab:datasets}.
For the sk-2005~\cite{BoVWFI}, twitter7~\cite{Kwak10www}, and wikipedia\_link\_en~\cite{konect} datasets, we have created them from existing real world graphs but with synthetic feature values just for the purpose of training time evaluation.
Datasets reddit~\cite{hamilton2017inductive}, ogbn-products, and ogbn-papers100M~\cite{hu2020open} are commonly used datasets in the field for comparing the training accuracies between different GNN architectures.
%

%

\textbf{Test System:} The platforms we have used for the evaluation are described in Table~\ref{tab:hardware}.
We use NVIDIA 450.51.05 driver and CUDA 10.2 on the evaluation platforms.
System2 and System3 configurations are only used in Section~\ref{sec.evaluation.microbenchmarkI}.

\begin{table}[t]
\caption{Evaluation Platforms}
\label{tab:hardware}
\vspace{-2.5mm}
\begin{center}
\begin{small}
\begin{tabular}{ccl}
\toprule
Config & Type & \multicolumn{1}{c}{Specification}\\
\midrule
System1 & CPU & AMD Threadripper 3960X 24C/48T\\
(Primary) & GPU & NVIDIA TITAN Xp 12GB\\
\midrule
\multirow{2}{*}{System2} & CPU & Dual Intel Xeon Gold 6230 40C/80T\\
& GPU & NVIDIA Tesla V100 16GB\\
\midrule
\multirow{2}{*}{System3} & CPU & Intel i7-8700K 6C/12T\\
& GPU & NVIDIA GTX 1660 6GB\\
\bottomrule
\end{tabular}
\end{small}
\end{center}
\end{table}


\textbf{Microbenchmark:} We would like to answer the following questions with the microbenchmark:

\begin{itemize}
    \item How does increasing the feature size affect the PyTorch-Direct performance? The feature sizes vary greatly across datasets. For example, while a node of ogbn-products~\cite{hu2020open} has 100 features, a node of reddit~\cite{hamilton2017inductive} has 602 features.
    \item How does increasing the number of features to be copied affect the PyTorch-Direct performance? Depending on  factors such as a
    connectivity of the input graph and the batch size, the number of
    neighboring nodes which need to be fetched per batch can vary.
    \item How well does the alignment optimization as discussed in Section~\ref{sec.PyTorch_direct.implementation.alignment} work with misaligned input features?
    \item What is the performance impact of using PyTorch-Direct on different systems?
\end{itemize}
The microbenchmark is designed to mimic the behavior of the data gathering and copy processes in the GNN training.
The microbenchmark uses a random number generator (RNG) to generate random indices which are used to index feature values.
The total number of items is fixed to 4M for all experiments.

\textbf{GNN Training:} In this evaluation, we use GraphSAGE~\cite{hamilton2017inductive} and graph attention network (GAT)~\cite{attention2018graph} implementations from the deep graph library (DGL).
Both implementations of DGL have all necessary supports (e.g. subgraph generation) to perform GNN mini-batching, which makes it suitable to work even if the input graphs cannot be fit in to the GPU memory.
The features are located in host memory and during training only the immediately required features are transferred to the GPU memory.
In the baseline implementation with PyTorch, the required features are gathered by the CPU and then copied to the GPU memory through DMA.
In the PyTorch-Direct implementation, the entire features are located in the unified tensor and the GPU directly accesses only the immediately required features.
Beside the data movement parts, the core training algorithms of the DGL implementations are left unmodified.

\subsection{Microbenchmark - Size and System Dependency}
\label{sec.evaluation.microbenchmarkI}

\begin{figure}[t]
  \centering
  \includegraphics[width=\linewidth]{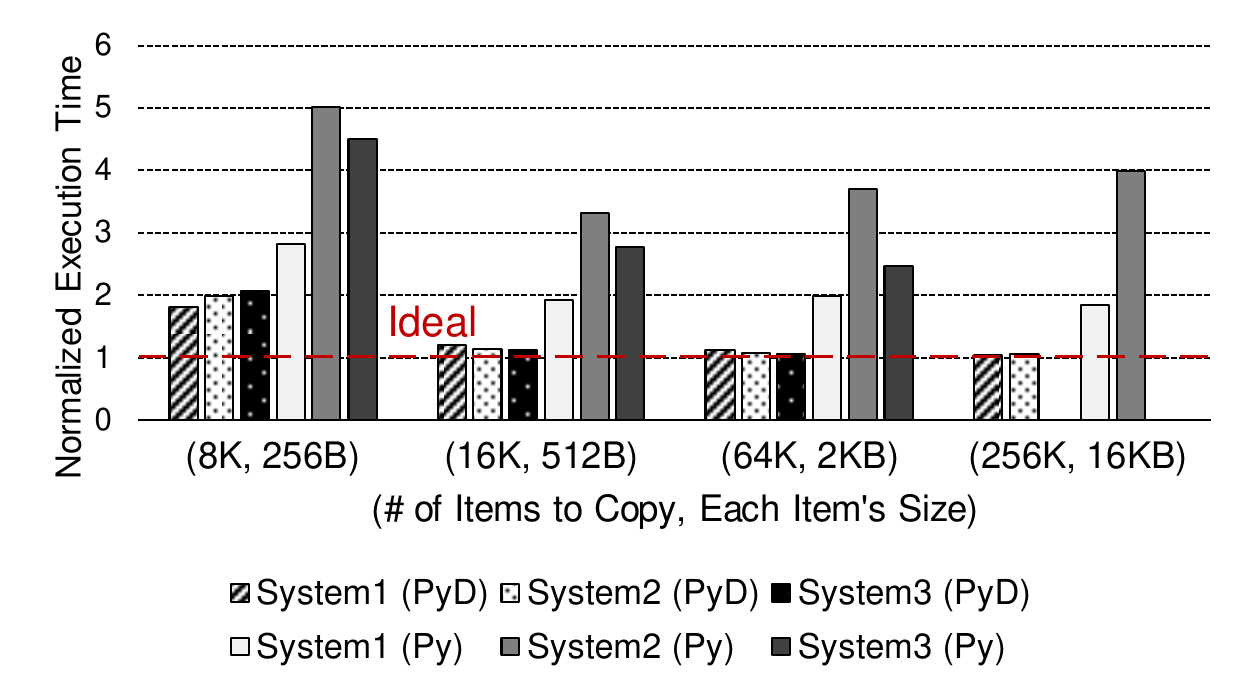}
  \caption{Irregular host data access pattern microbenchmark comparisons between PyTorch (Py) and PyTorch-Direct (PyD) on different systems. The ideal case shows only the pure data transfer time with a peak PCIe bandwidth.}
  \label{fig:micro_fig}
\end{figure}

The result of copying different numbers of features with different sizes are shown in Figure~\ref{fig:micro_fig}.
The ideal case only includes the pure data transfer time under the theoretical peak bandwidth of interconnect.
%
Due to the lack of system memory, we do not run the \texttt{(256K, 16KB)} setup with System3.
With the baseline PyTorch approach, the performance varies greatly depending on the system configurations.
While the slowdowns in System2 are about 3.31$\times$ to 5.01$\times$, and the slowdowns in System1 are about 1.85$\times$ to 2.82$\times$.
On the other hand, with PyTorch-Direct, we are able to consistently reach near to the ideal performance regardless of the system configuration unless the data transfer volume is very small.
When the total data transfer volume is very small, the overall execution time is dominated by the CUDA API calls and kernel launch overheads.
Except for the \texttt{(8K, 256B)} case, the baseline PyTorch approach shows 1.85$\times$ to 3.98$\times$ slowdowns while PyTorch-Direct shows only 1.03$\times$ to 1.20$\times$ slowdowns compared with the ideal case.
Overall, PyTorch-Direct shows about 2.39$\times$ of performance improvement in average compared to the baseline PyTorch approach.

%
%
%

\subsection{Microbenchmark - Memory Alignment}

\begin{figure}[t]
  \centering
  \includegraphics[width=\linewidth]{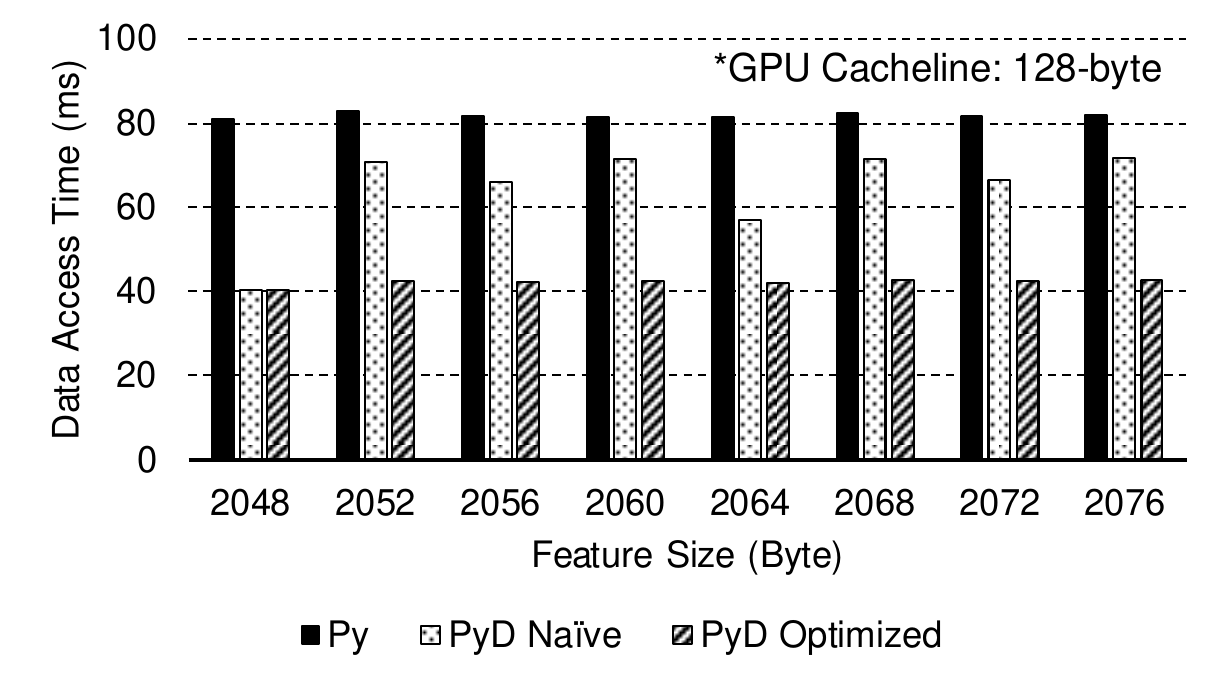}
  \caption{Memory access alignment and its impact on PyTorch-Direct (PyD) performance. PyTorch (Py) results added for comparison.}
  \label{fig:alignment}
  \vspace{-4mm}
\end{figure}

To evaluate the impact of the memory alignment optimization in PyTorch-Direct, we measure data access times for various feature sizes from 2048-byte to 2076-byte in 4-byte stride.
%
The evaluation result is shown in Figure~\ref{fig:alignment}.
For the \texttt{PyD Naïve} case, we use unmodified GPU indexing kernel from PyTorch and the kernel has no knowledge of memory alignment.
%
%
%
For the \texttt{PyD Optimized} case, the optimization from Section~\ref{sec.PyTorch_direct.implementation.alignment} is applied.

Figure~\ref{fig:alignment} shows that PyTorch-Direct reduces the data access time significantly compared to the PyTorch baseline.
However, the benefit is only limited without the memory-alignment optimization.
For example, when the feature size is 2052-byte, the \texttt{PyD Naïve}  provides only 1.17$\times$ of performance improvement over \texttt{Py}, while the \texttt{PyD Optimized} provides 1.95$\times$ of performance improvement.
Based on the results, we observe the optimization provides a consistent benefit over the PyTorch baseline (averagely 1.93$\times$) regardless of the data alignment.

\subsection{GNN Training Performance}
\label{sec.evaluation.gnn}


%
%
%
%

\begin{figure}[t]
  \centering
  \includegraphics[width=\linewidth]{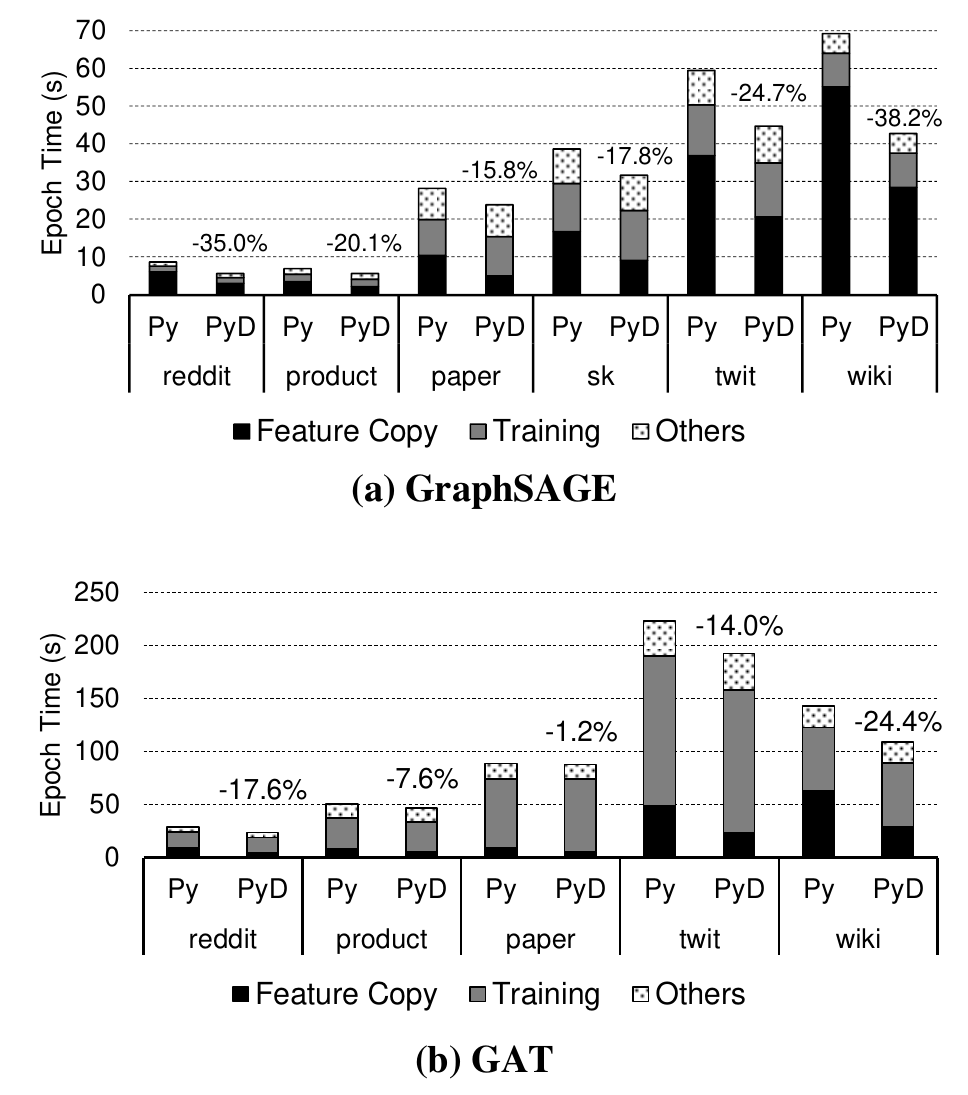}
  \caption{Single epoch execution time breakdown for both PyTorch (Py) vs. PyTorch-Direct (PyD) when running (a) GraphSAGE and (b) GAT in different datasets. Training epoch time reductions written on the bars.}
  \label{fig:graphsage}
\end{figure}

In Figure~\ref{fig:graphsage}, we compare the breakdown of the training epoch time when using unmodified DGL implementations in PyTorch vs. PyTorch-Direct.
In the GAT training, we do not run \texttt{sk} dataset due to the DGL's out-of-host-memory error for both PyTorch and PyTorch-Direct cases.
Similar to the microbenchmark results in Section~\ref{sec.evaluation.microbenchmarkI}, we observe about 47.1\% of reduction in the feature copy times.
The other portions of the training epoch times remain almost identical to the baseline case.
PyTorch-Direct gives less benefit for datasets with smaller feature sizes (e.g. \texttt{paper}), because the feature copy time is smaller in the end-to-end training time.
%
%
Similarly, GAT training is computationally heavier than GraphSAGE and therefore we observe a less benefit of PyTorch-Direct.
%
Overall, we observe between
1.01$\times$ to 1.45$\times$ speedup when we use PyTorch-Direct in GNN training.

\begin{figure}[t]
  \centering
  \includegraphics[width=\linewidth]{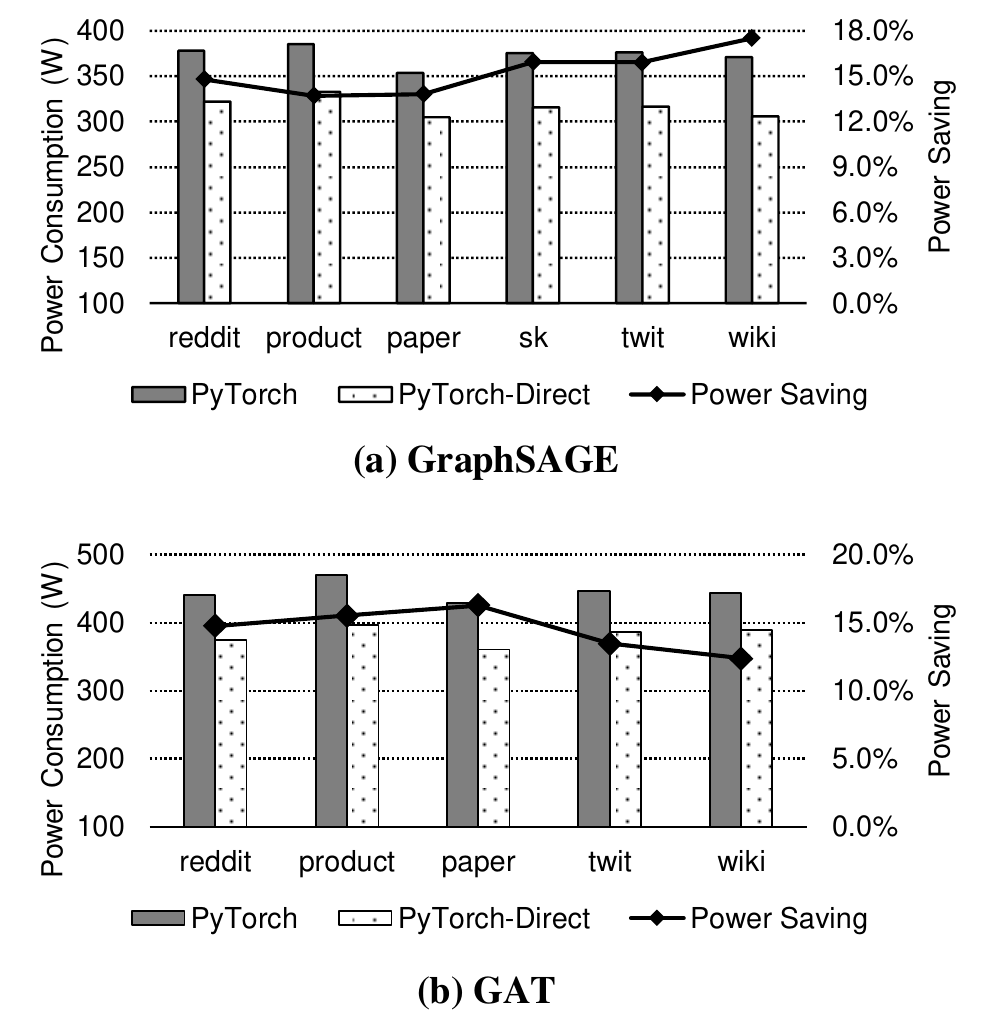}
  \caption{Total system power consumption comparison between PyTorch vs. PyTorch-Direct when running (a) GraphSAGE and (b) GAT in different datasets. System idle power is about 105W.}
  \label{fig:graphsage_power}
\end{figure}


With more efficient data access, in addition to the training speedup, PyTorch-Direct saves system power by 12.4\% to 17.5\% during GNN training.
%
Figure~\ref{fig:graphsage_power} compares power consumptions across different GNN trainings measured at electricity meter.
%

\section{Conclusion}
\label{sec.conclusion}

In this work we present PyTorch-Direct, which proposes a GPU-centric rather than a CPU-centric data access paradigm for GNN training.
%
We show that PyTorch-Direct is effective in reducing CPU utilization in GNN training and results in higher end-to-end training performance and lower power consumption.
%
For the input data sets and GNN architectures evaluated, PyTorch-Direct reduces the overall training time by up to 38.2\% and saves system power by up to 17.5\%.
PyTorch-Direct has minimal required programmer effort. Users can take full advantage of the benefits PyTorch-Direct provides by modifying at most two lines of their original code.
%

\nocite{langley00}

\bibliography{example_paper}

\begin{thebibliography}{28}
\providecommand{\natexlab}[1]{#1}
\providecommand{\url}[1]{\texttt{#1}}
\expandafter\ifx\csname urlstyle\endcsname\relax
  \providecommand{\doi}[1]{doi: #1}\else
  \providecommand{\doi}{doi: \begingroup \urlstyle{rm}\Url}\fi

\bibitem[Abadi et~al.(2015)Abadi, Agarwal, Barham, Brevdo, Chen, Citro,
  Corrado, Davis, Dean, Devin, Ghemawat, Goodfellow, Harp, Irving, Isard, Jia,
  Jozefowicz, Kaiser, Kudlur, Levenberg, Man\'{e}, Monga, Moore, Murray, Olah,
  Schuster, Shlens, Steiner, Sutskever, Talwar, Tucker, Vanhoucke, Vasudevan,
  Vi\'{e}gas, Vinyals, Warden, Wattenberg, Wicke, Yu, and
  Zheng]{tensorflow2015-whitepaper}
Abadi, M., Agarwal, A., Barham, P., Brevdo, E., Chen, Z., Citro, C., Corrado,
  G.~S., Davis, A., Dean, J., Devin, M., Ghemawat, S., Goodfellow, I., Harp,
  A., Irving, G., Isard, M., Jia, Y., Jozefowicz, R., Kaiser, L., Kudlur, M.,
  Levenberg, J., Man\'{e}, D., Monga, R., Moore, S., Murray, D., Olah, C.,
  Schuster, M., Shlens, J., Steiner, B., Sutskever, I., Talwar, K., Tucker, P.,
  Vanhoucke, V., Vasudevan, V., Vi\'{e}gas, F., Vinyals, O., Warden, P.,
  Wattenberg, M., Wicke, M., Yu, Y., and Zheng, X.
\newblock {TensorFlow}: Large-scale machine learning on heterogeneous systems,
  2015.
\newblock URL \url{https://www.tensorflow.org/}.
\newblock Software available from tensorflow.org.

\bibitem[Boldi \& Vigna(2004)Boldi and Vigna]{BoVWFI}
Boldi, P. and Vigna, S.
\newblock The {W}eb{G}raph framework {I}: {C}ompression techniques.
\newblock In \emph{Proceedings of the Thirteenth International World Wide Web
  Conference (WWW 2004)}, pp.\  595--601, Manhattan, USA, 2004. ACM Press.

\bibitem[Chiang et~al.(2019)Chiang, Liu, Si, Li, Bengio, and
  Hsieh]{chiang2019cluster}
Chiang, W.-L., Liu, X., Si, S., Li, Y., Bengio, S., and Hsieh, C.-J.
\newblock Cluster-gcn: An efficient algorithm for training deep and large graph
  convolutional networks.
\newblock In \emph{Proceedings of the 25th ACM SIGKDD International Conference
  on Knowledge Discovery \& Data Mining}, KDD '19, pp.\  257–266, New York,
  NY, USA, 2019. Association for Computing Machinery.
\newblock ISBN 9781450362016.
\newblock \doi{10.1145/3292500.3330925}.
\newblock URL \url{https://doi.org/10.1145/3292500.3330925}.

\bibitem[Data61(2018)]{StellarGraph}
Data61, C.
\newblock Stellargraph machine learning library.
\newblock \url{https://github.com/stellargraph/stellargraph}, 2018.

\bibitem[Fey \& Lenssen(2019)Fey and Lenssen]{Fey/Lenssen/2019}
Fey, M. and Lenssen, J.~E.
\newblock Fast graph representation learning with {PyTorch Geometric}.
\newblock In \emph{ICLR Workshop on Representation Learning on Graphs and
  Manifolds}, 2019.

\bibitem[Gera et~al.(2020)Gera, Kim, Sao, Kim, and Bader]{Gera20}
Gera, P., Kim, H., Sao, P., Kim, H., and Bader, D.
\newblock Traversing large graphs on gpus with unified memory.
\newblock \emph{Proceedings of the VLDB Endowment}, 13\penalty0 (7):\penalty0
  1119–1133, March 2020.

\bibitem[Grattarola \& Alippi(2020)Grattarola and Alippi]{grattarola2020graph}
Grattarola, D. and Alippi, C.
\newblock Graph neural networks in tensorflow and keras with spektral.
\newblock \emph{arXiv preprint arXiv:2006.12138}, 2020.

\bibitem[Hamilton et~al.(2017)Hamilton, Ying, and
  Leskovec]{hamilton2017inductive}
Hamilton, W.~L., Ying, R., and Leskovec, J.
\newblock Inductive representation learning on large graphs.
\newblock In \emph{Proceedings of the 31st International Conference on Neural
  Information Processing Systems}, NIPS'17, pp.\  1025–1035, Red Hook, NY,
  USA, 2017. Curran Associates Inc.
\newblock ISBN 9781510860964.

\bibitem[He et~al.(2015)He, Zhang, Ren, and Sun]{He2015}
He, K., Zhang, X., Ren, S., and Sun, J.
\newblock Deep residual learning for image recognition.
\newblock \emph{arXiv preprint arXiv:1512.03385}, 2015.

\bibitem[Hu et~al.(2020{\natexlab{a}})Hu, Fey, Zitnik, Dong, Ren, Liu, Catasta,
  and Leskovec]{hu2020ogb}
Hu, W., Fey, M., Zitnik, M., Dong, Y., Ren, H., Liu, B., Catasta, M., and
  Leskovec, J.
\newblock Open graph benchmark: Datasets for machine learning on graphs.
\newblock \emph{arXiv preprint arXiv:2005.00687}, 2020{\natexlab{a}}.

\bibitem[Hu et~al.(2020{\natexlab{b}})Hu, Fey, Zitnik, Dong, Ren, Liu, Catasta,
  and Leskovec]{hu2020open}
Hu, W., Fey, M., Zitnik, M., Dong, Y., Ren, H., Liu, B., Catasta, M., and
  Leskovec, J.
\newblock Open graph benchmark: Datasets for machine learning on graphs,
  2020{\natexlab{b}}.

\bibitem[Kipf \& Welling(2017)Kipf and Welling]{kipf2017semi}
Kipf, T.~N. and Welling, M.
\newblock Semi-supervised classification with graph convolutional networks.
\newblock In \emph{International Conference on Learning Representations
  (ICLR)}, 2017.

\bibitem[Krizhevsky et~al.(2012)Krizhevsky, Sutskever, and
  Hinton]{krizhevsky2012imagenet}
Krizhevsky, A., Sutskever, I., and Hinton, G.~E.
\newblock Imagenet classification with deep convolutional neural networks.
\newblock In \emph{Advances in neural information processing systems}, pp.\
  1097--1105, 2012.

\bibitem[Kunegis(2013)]{konect}
Kunegis, J.
\newblock Konect: The koblenz network collection.
\newblock In \emph{Proceedings of the 22nd International Conference on World
  Wide Web}, WWW '13 Companion, pp.\  1343–1350, New York, NY, USA, 2013.
  Association for Computing Machinery.
\newblock ISBN 9781450320382.
\newblock \doi{10.1145/2487788.2488173}.
\newblock URL \url{https://doi.org/10.1145/2487788.2488173}.

\bibitem[Kwak et~al.(2010)Kwak, Lee, Park, and Moon]{Kwak10www}
Kwak, H., Lee, C., Park, H., and Moon, S.
\newblock {W}hat is {T}witter, a social network or a news media?
\newblock In \emph{WWW '10: Proc. the 19th Intl. Conf. on World Wide Web}, pp.\
   591--600, New York, NY, USA, 2010. ACM.
\newblock ISBN 978-1-60558-799-8.
\newblock \doi{http://doi.acm.org/10.1145/1772690.1772751}.

\bibitem[Langley(2000)]{langley00}
Langley, P.
\newblock Crafting papers on machine learning.
\newblock In Langley, P. (ed.), \emph{Proceedings of the 17th International
  Conference on Machine Learning (ICML 2000)}, pp.\  1207--1216, Stanford, CA,
  2000. Morgan Kaufmann.

\bibitem[Liu et~al.(2020)Liu, Lu, Chen, and He]{liu2020g3}
Liu, H., Lu, S., Chen, X., and He, B.
\newblock G3: When graph neural networks meet parallel graph processing systems
  on gpus.
\newblock \emph{Proc. VLDB Endow.}, 13\penalty0 (12):\penalty0 2813–2816,
  August 2020.
\newblock ISSN 2150-8097.
\newblock \doi{10.14778/3415478.3415482}.
\newblock URL \url{https://doi.org/10.14778/3415478.3415482}.

\bibitem[Marcel \& Rodriguez(2010)Marcel and Rodriguez]{torchvision}
Marcel, S. and Rodriguez, Y.
\newblock {Torchvision the Machine-Vision Package of Torch}.
\newblock In \emph{Proceedings of the 18th ACM International Conference on
  Multimedia}, MM '10, pp.\  1485–1488, New York, NY, USA, 2010. Association
  for Computing Machinery.
\newblock ISBN 9781605589336.
\newblock \doi{10.1145/1873951.1874254}.
\newblock URL \url{https://doi.org/10.1145/1873951.1874254}.

\bibitem[Min et~al.(2020)Min, Mailthody, Qureshi, Xiong, Ebrahimi, and
  Hwu]{min2020emogi}
Min, S.~W., Mailthody, V.~S., Qureshi, Z., Xiong, J., Ebrahimi, E., and Hwu,
  W.-m.
\newblock Emogi: Efficient memory-access for out-of-memory graph-traversal in
  gpus.
\newblock \emph{arXiv preprint arXiv:2006.06890}, 2020.

\bibitem[NVIDIA(2016)]{uvmguide}
NVIDIA.
\newblock {Beyond GPU Memory Limits with Unified Memory on Pascal}, 2016.
\newblock URL
  \url{https://developer.nvidia.com/blog/beyond-gpu-memory-limits-unified-memory-pascal/}.

\bibitem[Paszke et~al.(2019)Paszke, Gross, Massa, Lerer, Bradbury, Chanan,
  Killeen, Lin, Gimelshein, Antiga, Desmaison, Kopf, Yang, DeVito, Raison,
  Tejani, Chilamkurthy, Steiner, Fang, Bai, and Chintala]{adam2019pytorch}
Paszke, A., Gross, S., Massa, F., Lerer, A., Bradbury, J., Chanan, G., Killeen,
  T., Lin, Z., Gimelshein, N., Antiga, L., Desmaison, A., Kopf, A., Yang, E.,
  DeVito, Z., Raison, M., Tejani, A., Chilamkurthy, S., Steiner, B., Fang, L.,
  Bai, J., and Chintala, S.
\newblock Pytorch: An imperative style, high-performance deep learning library.
\newblock In Wallach, H., Larochelle, H., Beygelzimer, A., d\' Alch\'{e}-Buc,
  F., Fox, E., and Garnett, R. (eds.), \emph{Advances in Neural Information
  Processing Systems 32}, pp.\  8024--8035. Curran Associates, Inc., 2019.

\bibitem[Sabet et~al.(2020)Sabet, Zhao, and Gupta]{Sabet20}
Sabet, A. H.~N., Zhao, Z., and Gupta, R.
\newblock Subway: Minimizing data transfer during out-of-gpu-memory graph
  processing.
\newblock In \emph{Proceedings of the Fifteenth European Conference on Computer
  Systems}, EuroSys ’20, New York, NY, USA, 2020. Association for Computing
  Machinery.

\bibitem[Veličković et~al.(2018)Veličković, Cucurull, Casanova, Romero,
  Liò, and Bengio]{attention2018graph}
Veličković, P., Cucurull, G., Casanova, A., Romero, A., Liò, P., and Bengio,
  Y.
\newblock Graph attention networks.
\newblock In \emph{International Conference on Learning Representations}, 2018.
\newblock URL \url{https://openreview.net/forum?id=rJXMpikCZ}.

\bibitem[Wang et~al.(2019)Wang, Zheng, Ye, Gan, Li, Song, Zhou, Ma, Yu, Gai,
  Xiao, He, Karypis, Li, and Zhang]{wang2019dgl}
Wang, M., Zheng, D., Ye, Z., Gan, Q., Li, M., Song, X., Zhou, J., Ma, C., Yu,
  L., Gai, Y., Xiao, T., He, T., Karypis, G., Li, J., and Zhang, Z.
\newblock Deep graph library: A graph-centric, highly-performant package for
  graph neural networks.
\newblock \emph{arXiv preprint arXiv:1909.01315}, 2019.

\bibitem[{Wu} et~al.(2020){Wu}, {Pan}, {Chen}, {Long}, {Zhang}, and
  {Yu}]{zu2019survey}
{Wu}, Z., {Pan}, S., {Chen}, F., {Long}, G., {Zhang}, C., and {Yu}, P.~S.
\newblock A comprehensive survey on graph neural networks.
\newblock \emph{IEEE Transactions on Neural Networks and Learning Systems},
  pp.\  1--21, 2020.

\bibitem[Ying et~al.(2018)Ying, He, Chen, Eksombatchai, Hamilton, and
  Leskovec]{ying2019pinsage}
Ying, R., He, R., Chen, K., Eksombatchai, P., Hamilton, W.~L., and Leskovec, J.
\newblock Graph convolutional neural networks for web-scale recommender
  systems.
\newblock In \emph{Proceedings of the 24th ACM SIGKDD International Conference
  on Knowledge Discovery \& Data Mining}, KDD '18, pp.\  974–983, New York,
  NY, USA, 2018. Association for Computing Machinery.
\newblock ISBN 9781450355520.
\newblock \doi{10.1145/3219819.3219890}.
\newblock URL \url{https://doi.org/10.1145/3219819.3219890}.

\bibitem[Zeng et~al.(2020)Zeng, Zhou, Srivastava, Kannan, and
  Prasanna]{graphsaint-iclr20}
Zeng, H., Zhou, H., Srivastava, A., Kannan, R., and Prasanna, V.
\newblock {GraphSAINT}: Graph sampling based inductive learning method.
\newblock In \emph{International Conference on Learning Representations}, 2020.
\newblock URL \url{https://openreview.net/forum?id=BJe8pkHFwS}.

\bibitem[Zhou et~al.(2018)Zhou, Cui, Zhang, Yang, Liu, and Sun]{zhou2019gnn}
Zhou, J., Cui, G., Zhang, Z., Yang, C., Liu, Z., and Sun, M.
\newblock Graph neural networks: {A} review of methods and applications.
\newblock \emph{CoRR}, abs/1812.08434, 2018.
\newblock URL \url{http://arxiv.org/abs/1812.08434}.

\end{thebibliography}
\bibliographystyle{mlsys2020}



\end{document}